\documentclass{article}

\PassOptionsToPackage{numbers, compress}{natbib}
\usepackage[preprint]{neurips_2026}

\usepackage[utf8]{inputenc} 
\usepackage[T1]{fontenc}    
\usepackage{url}            
\usepackage{booktabs}       
\usepackage{amsfonts}       
\usepackage{nicefrac}       
\usepackage{microtype}      

\usepackage{graphicx}
\usepackage{amsmath}
\usepackage{amssymb}
\usepackage{caption}
\usepackage{verbatimbox}
\usepackage{diagbox}
\usepackage{multicol}
\usepackage{enumerate}
\usepackage{epsfig}
\usepackage{threeparttable}
\usepackage{enumitem}
\usepackage{multirow}
\usepackage{color}
\usepackage{array}
\usepackage{setspace}
\usepackage{makecell}
\usepackage{wrapfig}
\usepackage[accsupp]{axessibility} 

\usepackage[dvipsnames]{xcolor}
\definecolor{c1}{HTML}{5593ff}
\definecolor{c2}{HTML}{e83100}

\usepackage[colorlinks,
linkcolor=c2,       
anchorcolor=blue,  
citecolor=c1,        
]{hyperref}

\usepackage{xcolor}         

\title{ST-Gen4D: Embedding 4D Spatiotemporal Cognition \\ into World Model for 4D Generation}

\author{%
	Haonan Wang$\mathbf{^{1}}$, Hanyu Zhou$\mathbf{^{2}}$\thanks{Corresponding author.} , Tao Gu$\mathbf{^{3}}$, Luxin Yan$\mathbf{^{1}}$\\
	$\mathbf{^{1}}$School of Artificial Intelligence and Automation, Huazhong University of Science and Technology\\
	$\mathbf{^{2}}$School of Computing, National University of Singapore \\
	$\mathbf{^{3}}$School of Computing, Macquarie University \\
	\texttt{\{whn\_aurora,yanluxin\}@hust.edu.cn, hy.zhou@nus.edu.sg, tao.gu@mq.edu.au} \\
}

\begin{document}

\maketitle

\begin{abstract}
  Generative models have achieved success in producing apparently coherent 2D videos, but remain challenging in the physical world due to lack of 4D spatiotemporal scale. Typically, existing 4D generative models directly embed macro scale constraints to enhance overall spatiotemporal consistency. However, these methods only ensure global appearance coherence and fail to reveal the local dynamics of the physical world. Our insight is that global appearance structure and local dynamic topology empower 4D spatiotemporal cognition, thereby enabling 4D generation with spatiotemporal regularities. In this work, we propose ST-Gen4D, a 4D generation framework with 4D spatiotemporal cognition-based world model. Our model is guided by four key designs: 1) Spatiotemporal representation. We encode various modalities into multiple representations as a feature basis. 2) Spatiotemporal cognition. We sculpture these representations into global appearance graph and local dynamic graph, and fuse them via semantic-bridged spatiotemporal fusion to obtain a 4D cognition graph. 3) Spatiotemporal reasoning. We utilize a world model to derive future state based on the 4D cognition. 4) Spatiotemporal generation. We leverage the derived cognition as condition to guide latent diffusion for 4D Gaussian generation. By deeply integrating 4D intrinsic cognition with generative priors, our model guarantees the structural rationality and topological consistency of 4D generation. Moreover, we propose ST-4D datasets by aggregating public 4D datasets and self-built subset. Extensive experiments demonstrate the superiority of our ST-Gen4D across 3D and 4D generation tasks.
\end{abstract}

\section{Introduction}
\label{intro}

Generative models have achieved remarkable progress in 2D video generation by learning the distribution mapping between textual prompts and spatiotemporal visual signals \cite{ho2022imagen,singer2022make,blattmann2023align,hong2022cogvideo,villegas2022phenaki,brooks2024video}. However, lifting such generative capability to the 4D physical world remains challenging, since 2D generative priors mainly capture frame-level appearance evolution but lack explicit modeling of 4D spatiotemporal scale. In the physical world, dynamic scenes are continuous 4D processes governed by spatial-temporal regularities rather than independent visual frames. Therefore, effective 4D generation requires a deeper spatiotemporal understanding to support coherent generation across both space and time.

Existing 4D generation methods typically introduce macro-scale constraints, such as 4D-aware diffusion priors \cite{jiang2023consistent4d,zhao2023animate124,zhu2025worldsplat} or smoothness regularizations \cite{ling2024align,bahmani20244d,lyu2026choreographing}, to enhance overall spatiotemporal consistency. Although these constraints improve global coherence, they are usually imposed externally and mainly regularize generated results at a coarse level. Thus, they remain insufficient to reveal the intrinsic organization of dynamic 4D scenes, especially how overall appearance structures are preserved and how local regions move, deform, and interact over time. This motivates us to move beyond macro-scale constraints and introduce an intrinsic spatiotemporal cognition for 4D generation.

\begin{figure}[t]
	\centering
	\setlength{\abovecaptionskip}{0.1cm} 
	\setlength{\belowcaptionskip}{-0.7cm} 
	\includegraphics[scale=0.365]{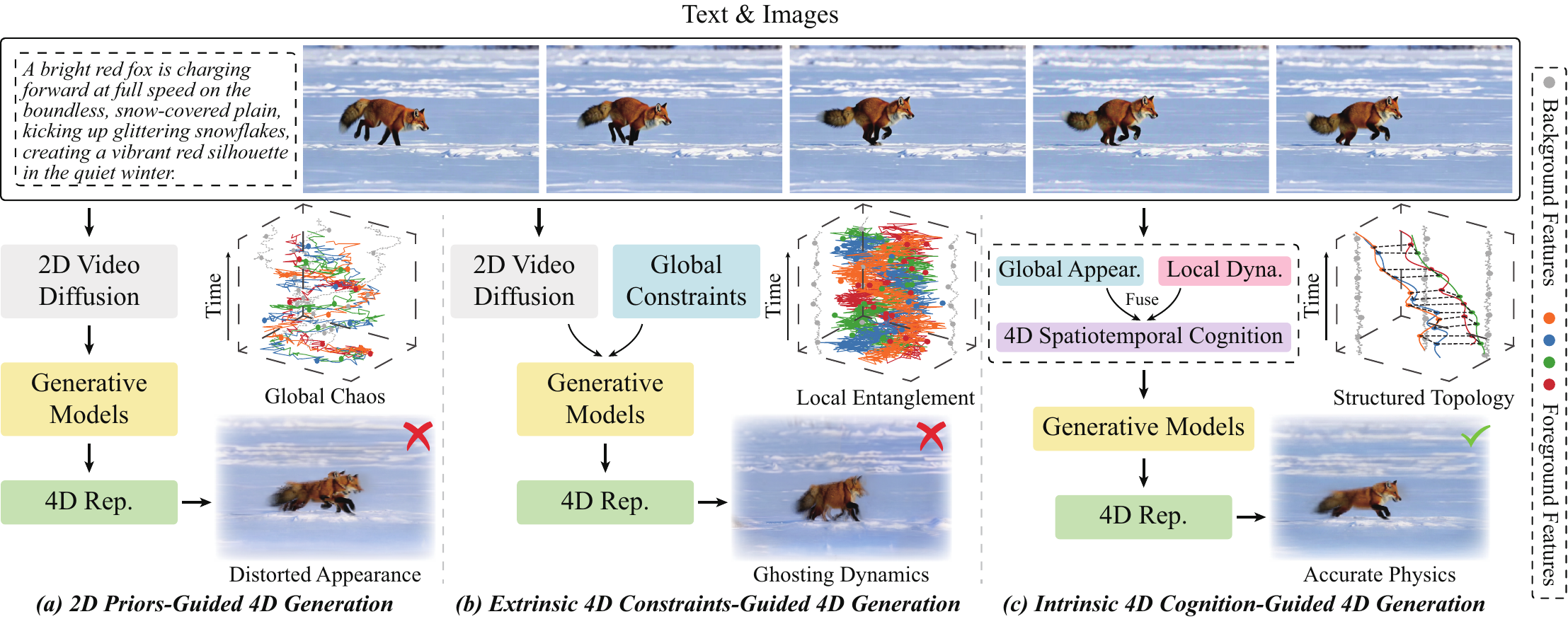}
	\caption{\textbf{Paradigm shift of 4D generation.} 2D Priors-guided generation mainly relies on 2D video diffusion, leading to collapsed appearance. Extrinsic constraints-guided generation introduces macro scale constraints but suffers from motion artifacts. Our intrinsic 4D cognition-guided paradigm leverages a spatiotemporal cognition to ensure plausible appearance and coherent dynamics.}
	\label{fig1}
\end{figure}

In this work, we define 4D spatiotemporal cognition as a structured state that encodes global appearance structure and local dynamic topology. The former describes spatial organization and cross-view consistency, while the latter characterizes temporal evolution and cross-time coherence. As shown in Fig. \ref{fig1}, this design is driven by two key insights: 1) Spatiotemporal encoders capture deep visual correlations to support robust spatial alignment and cross-time continuity. 2) Cognition graphs impose structural and topological constraints to preserve semantic boundaries and mitigate feature entanglement during complex object interactions. More importantly, unlike static conditions or external regularization terms, our cognition can be evolved by a world model to infer future spatiotemporal states, which are further used to condition the subsequent 4D generation process.

Based on this idea, we propose \textbf{ST-Gen4D}, a 4D generation framework with a spatiotemporal cognition-based world model. Our framework consists of four stages: 1) Spatiotemporal foundational representation. We introduce the 4D-VGGT \cite{wang20254d} encoder to extract robust features for cross-view consistency and temporal continuity. 2) Spatiotemporal structural cognition. We construct a global appearance graph and a local dynamic graph, and fuse them to obtain spatiotemporal cognition. 3) Spatiotemporal predictive reasoning. We utilize the world model to derive the evolution of dynamic scenes based on cognition. 4) Spatiotemporal consistent generation. We leverage the predicted cognition as the condition of a latent diffusion for 4D Gaussians generation, ensuring the appearance rationality and dynamics consistency. By integrating intrinsic 4D cognition with generative priors, ST-Gen4D achieves spatially stable and temporally coherent 4D generation.

Our main contributions are summarized as:
\begin{itemize}[leftmargin=*]
	\item We propose ST-Gen4D, a novel 4D generation framework with 4D spatiotemporal cognition-based world model. This cognitive guidance allows our model to maintain superior spatial stability and temporal coherence for physical world.
	\item Observing that spatiotemporal encoders ensure robust spatiotemporal consistency while cognition graphs provide structural constraints, we design a world model reasoning paradigm based on 4D cognition to accurately predict dynamic scene evolution.
	\item We develop a latent diffusion generation process guided by the future 4D spatiotemporal cognition inferred from the world model. Conditioning 4D Gaussian generation on these predicted states enforces appearance structure and dynamic topologic consistency.
	\item We aggregate a series of 4D datasets and build a validation subset, collectively named the spatiotemporal 4D dataset (\textbf{ST-4D}). Extensive experiments have demonstrated our significant effectiveness.
\end{itemize}

\section{Related Works}
\label{related_works}

\textbf{4D Generation.}
Driven by the success of 2D video generation, recent efforts in 4D generation predominantly utilize 2D video diffusion models to optimize or reconstruct 4D scenes \cite{jiang2023consistent4d,zhao2023animate124,zhu2025worldsplat,bahmani20244d,lyu2026choreographing,bahmani2025lyra,zheng2024unified,liu2025free4d}. For instance, Consistent4D \cite{jiang2023consistent4d} optimizes 4D scenes via SDS loss \cite{poole2022dreamfusion}, whereas Lyra \cite{bahmani2025lyra} directly reconstructs them from generated videos. Despite visual appeal, their reliance on 2D priors lacks 4D spatiotemporal scale, resulting in spatial structural distortions and temporal incoherence. Consequently, our work aims to introduce intrinsic 4D spatiotemporal modeling to significantly enhance spatiotemporal representation capabilities.

\begin{figure}[t]
	\centering
	\setlength{\abovecaptionskip}{0.1cm} 
	\setlength{\belowcaptionskip}{-0.7cm} 
	\includegraphics[scale=0.308]{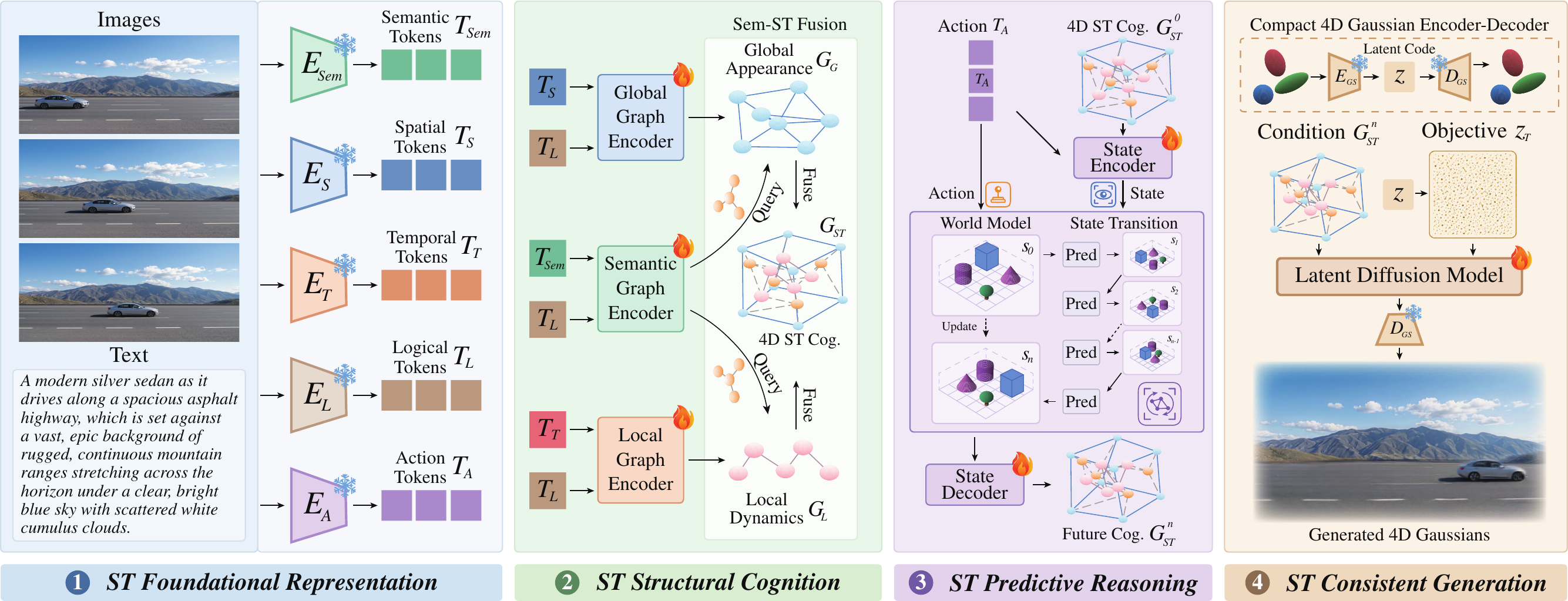}
	\caption{\textbf{Overview of ST-Gen4D.} The model first obtains multiple cognition tokens, which are then encoded into structured spatiotemporal cognition graph. We then utilize an autoregressive world model to predict state transition based on the cognition and action tokens. Finally, the deduced state severs as the condition of latent diffusion to generate 4D Gaussians.}
	\label{fig2}
\end{figure}

\textbf{Spatiotemporal-based 4D Generation.}
Existing 4D generation methods typically introduce macro-scale constraints like 4D-aware diffusion \cite{jiang2023consistent4d,zhao2023animate124,zhu2025worldsplat} or smoothness regularizations \cite{ling2024align,bahmani20244d,lyu2026choreographing} to enhance overall spatiotemporal consistency. Although these approaches improve macroscopic appearance consistency, they struggle to accurately model local dynamics in the physical world. This is because applying uniform global regularizations causes spatial-temporal over-smoothing, thereby ignoring complex dynamics. Thus, we introduce a structured spatiotemporal cognition graph to simultaneously comprehend global appearance structures and local dynamic topologies.

\textbf{World Model.}
Achieving 4D generation with plausible appearance and coherent dynamics relies on the accurate spatiotemporal deduction of the physical world. While world models \cite{lecun2022path,bardes2024revisiting,maes2026leworldmodel} excel at predicting these transitions, existing frameworks restrict them to inferring superficial pixel-level appearances across views and times \cite{lu2025gwm,chen2025teleworld,yang2026neoverse,team2026inspatio}. This neglects their intrinsic capacity to predict future structural relationships. Consquently, we aim to leverage the world model to simulate the transition process of 4D spatiotemporal cognition, thus accurately predicting the appearance structure and dynamic topology evolution of physical world.

\section{Our ST-Gen4D}
\label{sec3}
\textbf{Overview.} As shown in Fig. \ref{fig2}, our ST-Gen4D achieves structurally-rational and topologically-consistent 4D generation through the following four stages:

\noindent
1) \textbf{Spatiotemporal Foundational Representation.} This is the foundational representaion stage that we extract multiple tokens ($T_{Sem}, T_{S}, T_{T}, T_L, T_A$) from input image sequences $I$ and texts $T$ to serve as fundamental building blocks:
\begin{equation}
	\small
	\setlength\abovedisplayskip{0pt}
	\setlength\belowdisplayskip{-4pt}
	\begin{aligned}
		T_{Sem}, T_{S}, T_{T}, T_L, T_A = E_{Found}(I,T).
	\end{aligned}
\end{equation}

\noindent
2) \textbf{Spatiotemporal Structural Cognition.} This is the cognition construction stage that we encode tokens into independent topological graphs and fuse them via semantic-bridged spatiotemporal fusion (Sem-STF) to yield a spatiotemporal cognition graph $G_{ST}$:
\begin{equation}\small
	\setlength\abovedisplayskip{0pt}
	\setlength\belowdisplayskip{-4pt}
	\begin{aligned}
		G_{Sem},G_{S},G_{T}=\mathrm{GE}(T_{Sem,S,T}, T_L), G_{ST}=\mathrm{Sem-STF}(G_{Sem},G_{S},G_{T}).
	\end{aligned}
\end{equation}

\noindent
3) \textbf{Spatiotemporal Predictive Reasoning.} This is the state reasoning stage that we apply an autoregressive world model that leverages the initial cognition graph $G_{ST}^{t}$ and action tokens $T_A$ to infer a continuous sequence of future cognitive states $\hat{G}_{ST}^{t+1:T}$ with strict kinematic priors:
\begin{equation}\small
	\setlength\abovedisplayskip{0pt}
	\setlength\belowdisplayskip{-4pt}
	\begin{aligned}
		\hat{G}_{stc}^{t+1:T} = WM(G_{ST}^{t}, T_A).
	\end{aligned}
\end{equation}

\noindent
4) \textbf{Spatiotemporal Consistent Generation.} This is the cognition-guided generation stage that the predicted cognition steers a latent diffusion process, which is decoded into 4D Gaussians $\hat{\mathcal{G}}_{4D}$:
\begin{equation}
	\small
	\setlength\abovedisplayskip{0pt}
	\setlength\belowdisplayskip{-4pt}
	\begin{aligned}
		Z_{4D}^{(0)} = \text{CG-DiT}(Z_{4D}^{(T)} \mid \hat{G}_{ST}^{t+1:T}), \quad \hat{\mathcal{G}}_{4D} = D_{GS}(Z_{4D}^{(0)}).
	\end{aligned}
\end{equation}

\noindent
\textbf{Remarks.} The output $\hat{\mathcal{G}}_{4D}$ denotes the generated 4D Gaussians representing the dynamic scene. Ultimately, our framework leverages spatiotemporal cognition graphs as structural conditions to enable structurally-rational and topologically-consistent 4D generation in diffusion models.

\subsection{Spatiotemporal Foundational Representation}
To establish a robust feature foundation for subsequent cognition construction and world model reasoning, we propose a foundational representation framework consisting of semantic, spatial, temporal, logical, and action encoders to extract distinct representation.

\begin{wrapfigure}{r}{0.63\columnwidth} 
	\setlength{\abovecaptionskip}{3pt}
	\setlength{\belowcaptionskip}{-10pt}
	\vspace{-10pt} 
	\centering
	\includegraphics[width=0.63\columnwidth]{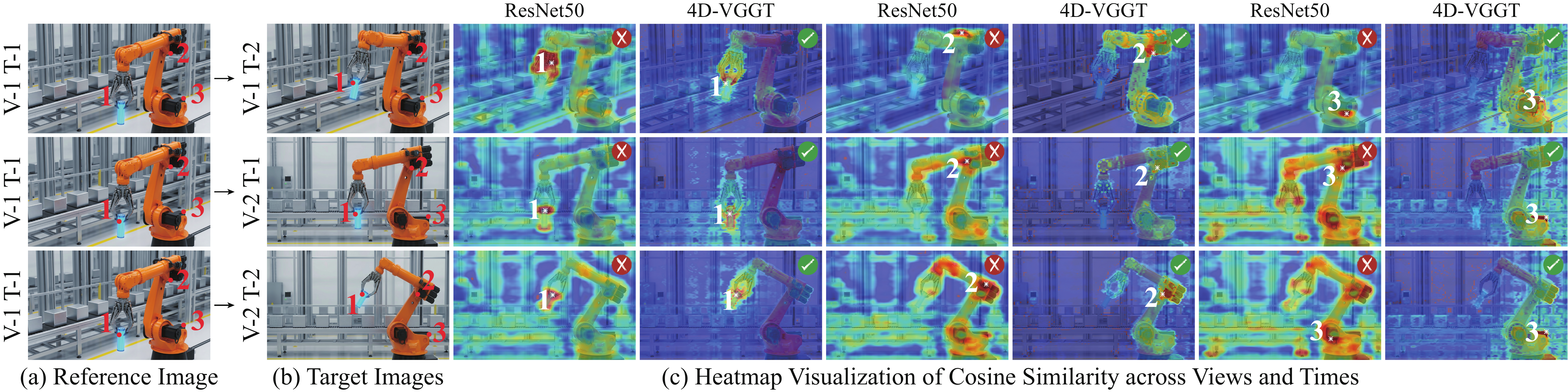}
	\caption{
		Cross-view and cross-time consistency analysis.
	} 
	\label{fig3}
\end{wrapfigure}

\textbf{Spatial and Temporal Encoders.} 
To establish robust 4D cognition, we extract visual semantics, global appearance, and local dynamics. We employ the Wan2.1 3D Causal VAE \cite{wan2025wan} for semantic tokens extraction. As Fig. \ref{fig3} shows, 4D-VGGT \cite{wang20254d} overcomes the feature drift of ResNet50 \cite{he2016deep}, providing strict cross-view and cross-time consistency. Therefore, we adopt 4D-VGGT as core encoders. Sharing a DINOv2 backbone for feature extraction, the spatial encoder integrates multi-view appearance via pair-wise cross-attention, while the temporal encoder captures local dynamics using sliding window attention \cite{wang20254d}, paving the way for structural cognition.

\textbf{Logical and Action Encoders.} 
To bridge spatiotemporal cognition and provide latent dynamic cues for reasoning, we introduce logical and action encoders. The logic encoder extracts image and text features via CLIP ViT and Text encoders \cite{radford2021learning} before mapping them through an MLP into unified logical tokens. Concurrently, the action encoder leverages InternVideo2 \cite{wang2024internvideo2} and a multimodal alignment strategy to distill inputs into tokens enriched with action priors. Ultimately, logical representations solidify cognition while action priors facilitate spatiotemporal inference.

\begin{wrapfigure}{r}{0.62\columnwidth} 
	\setlength{\abovecaptionskip}{3pt}
	\setlength{\belowcaptionskip}{-10pt}
	\vspace{-10pt} 
	\centering
	\includegraphics[width=0.62\columnwidth]{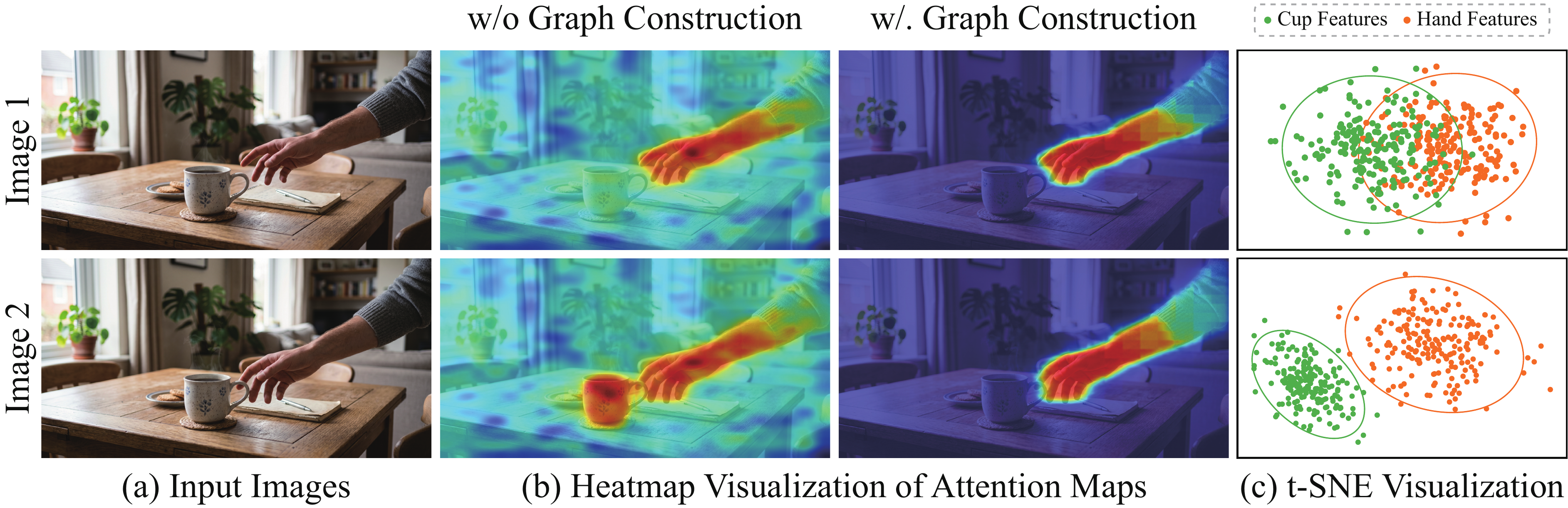}
	\caption{
		Feature robustness and discriminability analysis.
	} 
	\label{fig4}
\end{wrapfigure}

\subsection{Spatiotemporal Structural Cognition}

\begin{figure}[t]
	\centering
	\setlength{\abovecaptionskip}{0.1cm} 
	\setlength{\belowcaptionskip}{-0.7cm} 
	\includegraphics[scale=0.56]{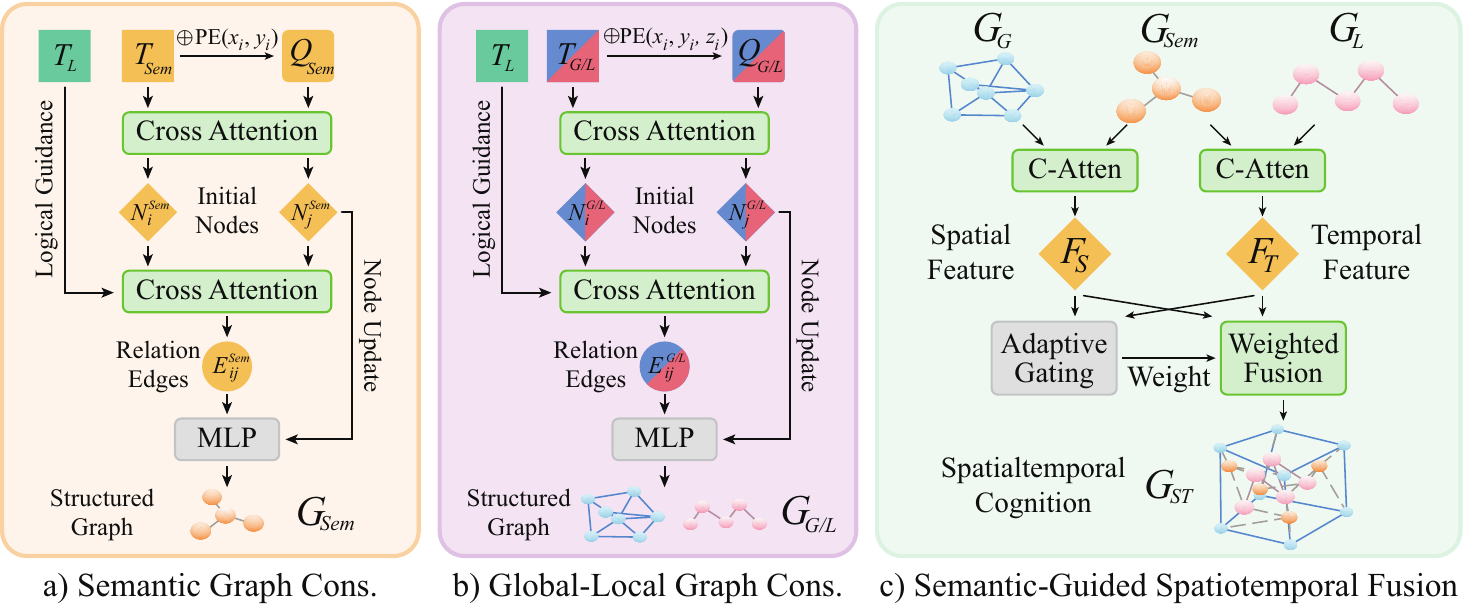}
	\caption{\textbf{Architectural details of the spatiotemporal structured cognition.} (a, b) The construction pipelines for semantic, global, and local graphs from foundational representations. (c) Details of the fusion progress between global and local graphs guided by common semantic graph.}
	\label{fig5}
\end{figure}

High-fidelity 4D generation requires guidance equipped with global appearance and local dynamics. As Fig. \ref{fig4} demonstrates, simple semantic features often fail to distinguish distinct entities during complex interactions. On the contrary, structured spatiotemporal features maintain high discriminability, driving our design of a structured spatiotemporal cognition in Fig. \ref{fig5} for robust inference.

\textbf{Semantic-Global-Local Graph Encoder.}
To capture comprehensive visual cues, we construct three graphs for semantics, global appearance, and local dynamics, all sharing a unified encoding paradigm.

\emph{Semantic Graph Construction.} We formulate the semantic query $Q_{Sem}$ by concatenating tokens $T_{Sem}$ with a 2D positional embedding $\text{PE}(x_p, y_p)$. Cross-attention between $Q_{Sem}$ and $T_S{Sem}$ yields $n$ initial nodes $N_i^{Sem}$. Guided by logical tokens $T_L$, we then establish semantic edges $E_{ij}^{Sem}$ between nodes $N_i^{Sem}$ and $N_j^{Sem}$. Finally, an MLP aggregates the resulting messages $m_{ij}^{Sem}$ to update the nodes, forming the semantic graph $G_{Sem}$:
\begin{equation}\small
	\small
	\setlength\abovedisplayskip{0pt}
	\setlength\belowdisplayskip{-6pt}
	\begin{aligned}
		m_{ij}^{Sem}=\text{MLP}(N_i^{Sem},N_j^{Sem},E_{ij}^{Sem}),\quad G_{sem}=N_i^{Sem}+\sum_{j}m_{ij}.
	\end{aligned}
\end{equation}

\emph{Global Appearance and Local Dynamic Graph Construction.} We construct the global appearance graph $G_G$ and local dynamics graph $G_L$ following a similar procedure, substituting semantic tokens with spatial and temporal tokens, respectively. To incorporate 3D geometric awareness, we replace the 2D positional embedding with a 3D geometry embedding $\text{GE}(x_p, y_p, z_p)$.

\textbf{Semantic-Bridged Spatiotemporal Fusion.} To aggregate these isolated graphs, we design a semantic-bridged spatiotemporal fusion module. Treating the semantic graph $G_{Sem}$ as a query, we apply dual-stream cross-attention to the global appearance graph $G_G$ and local dynamics graph $G_L$ to extract aligned features. An adaptive gating mechanism then generates dynamic weights, selectively aggregating these spatial and temporal streams into the semantic anchor via a feed-forward network. The resulting spatiotemporal cognition graph $G_{ST}$ provides a robust state for subsequent inference.

\subsection{Spatiotemporal Predictive Reasoning}
Building upon the spatiotemporal cognition graph and the extracted action tokens, this stage aims to extrapolate the current state into continuous future dynamics. Therefore, we propose a spatiotemporal predictive reasoning framework driven by a world model based on spatiotemporal cognition.

\textbf{World Action-State Formulation.} 
We distill the expansive cognition graph $G_{ST}$ into a compact state $s_t$ via cross-attention with learnable queries. Concurrently, action tokens $T_A$ are processed through temporal attention pooling and an MLP to form a dynamic condition vector $a_t$. Using adaptive layer normalization (adaLN), $a_t$ generates affine parameters to modulate $s_t$ into an action-conditioned state $s_t^{'}$. This integration enforces kinematic constraints during autoregressive modeling.

\textbf{Cognition-based World Model Reasoning.} 
To deduce future cognition, we instantiate a causal Transformer predictor inspired by LeWorldModel \cite{maes2026leworldmodel}. It processes the historical sequence $[s_{t-T}^{'}, \dots, s_t^{'}]$ to infer the subsequent latent state $\hat{s}_{t+1}$. A cross-attention state resampling decoder then projects this prediction into the cognition space, utilizing the original graph $G_{ST}^{t}$ as queries and $\hat{s}_{t+1}$ as keys and values to iteratively construct the future cognition $\hat{G}_{ST}^{t+1}$. This reasoning ensures future states preserve rigorous spatiotemporal dynamics, providing physically plausible guidance for 4D generation.

\subsection{Spatiotemporal Consistent Generation}
Having established a robust sequence of predicted spatiotemporal cognition graphs, the final objective is to generate high-fidelity 4D scenes. Therefore, we propose cognition-guided latent diffusion model and a latent 4D Gaussian encoder-decoder for generating physical-consistent 4D scenes.

\textbf{Cognition-Guided Latent Diffusion.} 
This module generates 4D scene latents by conditioning diffusion on the predicted cognition graphs. We employ a cognition-guided diffusion Transformer (CG-DiT) optimized via flow-matching. Stacked DiT blocks process an initial noise sequence $z_{4D}^{(T)}$, modulated by timestep $\tau$ via adaLN. Crucially, a structured cross-attention layer conditions intermediate features using the predicted graph $\hat{G}_{ST}$ as keys and values. A projection head then progressively denoises the sequence into the clean latent $z_{4D}^{(0)}$, ensuring the generated scenes maintain plausible spatial appearances and coherent temporal dynamics.

\textbf{Latent 4D Gaussian Encoder-Decoder.}
To facilitate latent generation, we compress massive 4D Gaussians into a tractable sequence using a pre-trained KL-regularized autoencoder. A sparse 4D convolutional encoder $E_{GS}$ voxelizes Gaussian points into a compact latent representation $z_{4D}$. Then a lightweight decoder $D_{GS}$ with self-attention and parallel MLPs reconstructs spatial attributes and polynomial deformation fields from these tokens. This mitigates the prohibitive overhead of generating millions of Gaussian splats, bridging explicit 4D rendering and diffusion generation.

\subsection{Optimization}
To effectively train our ST-Gen4D framework across the progressive stages, we optimize the system using the following three loss functions:

\textbf{World Model Loss ($\mathcal{L}_{WM}$).} The world model is optimized to learn accurate state transitions while maintaining a robust latent space. This loss consists of a MSE prediction loss and a SIGReg term for preventing representation collapse\cite{maes2026leworldmodel}:
\begin{equation}
	\small
	\setlength\abovedisplayskip{0pt}
	\setlength\belowdisplayskip{-4pt}
	\label{eq6}
	\mathcal{L}_{WM} = \mathbb{E}_{t} \left[ |\hat{s}_{t+1} - s_{t+1}|_2^2 \right] + \alpha \mathcal{L}_{SIGReg}(s),
\end{equation}

where $z_{t+1}$ and $\hat{z}_{t+1}$ represent the ground-truth and predicted future latent states, respectively, and $\alpha$ is the regularization weight.

\textbf{SDS Loss ($\mathcal{L}_{SDS}$).} To supervise the appearance and leverage generative priors, we employ the Score Distillation Sampling (SDS) loss from a pre-trained multi-view or video diffusion model:
\begin{equation}
	\small
	\setlength\abovedisplayskip{0pt}
	\setlength\belowdisplayskip{-1pt}
	\mathcal{L}_{SDS}(\phi,x) = \mathbb{E}_{t, \epsilon, c} \left[ w(t) \|\hat{\epsilon}_{\phi}(x_t; t, c) - \epsilon\|_2^2 \right],
\end{equation}
where $\epsilon$ is our predicted noise and $\hat{\epsilon}_{\phi}$ denotes the frozen pre-trained noise predictor. $x_t$ is the noisy latent at diffusion timestep $t$, and $c$ represents the images and texts conditioning prompts.

\textbf{Spatiotemporal Loss ($\mathcal{L}_{ST}$):} To equip the model with spatiotemporal perception, we introduce a spatiotemporal loss comprising Chamfer Distance (CD) for spatial supervision, temporal smoothness for deformation consistency, and a DDIM-based reconstruction loss for rendering fidelity:
\begin{equation}
	\small
	\setlength\abovedisplayskip{0pt}
	\setlength\belowdisplayskip{-4pt}
	\label{eq8}
	 \mathcal{L}_{ST} = \mathcal{L}_{CD}(\hat{\mathcal{G}}, \mathcal{G}) + \beta \sum_{t} \| \Delta \hat{\mathcal{G}}_{t} - \Delta \hat{\mathcal{G}}_{t-1} \|_2^2 + \gamma \mathcal{L}_{DDIM}(I(\hat{\mathcal{G}}), I(\mathcal{G})),
\end{equation}
where $\hat{\mathcal{G}}$ and $\mathcal{G}$ are the generated and ground-truth 4D Gaussians, $\Delta \hat{\mathcal{G}}_{t}$ denotes vertex displacement between adjacent Gaussians, $I(\cdot)$ represents rendered images, and $\beta, \gamma$ are weighting factors.

The total optimization objective is formulated as:
\begin{equation}
	\small
	\setlength\abovedisplayskip{0pt}
	\setlength\belowdisplayskip{-1pt}
	\label{eq9}
	\mathcal{L}_{total} = \lambda_{1} \mathcal{L}_{WM} + \lambda_{2} \mathcal{L}_{SDS} + \lambda_{3} \mathcal{L}_{ST},
\end{equation}
where $\lambda{1}$, $\lambda_{2}$, and $\lambda_{3}$ are hyperparameters balancing the contributions of state prediction, appearance alignment, and spatiotemporal refinement.

\section{Training Datasets and Pipeline}
\label{sec4}

\subsection{Our ST-4D Dataset}

\begin{wrapfigure}{r}{0.62\columnwidth} 
	\setlength{\abovecaptionskip}{3pt}
	\setlength{\belowcaptionskip}{-5pt}
	\vspace{-10pt} 
	\centering
	\includegraphics[width=0.62\columnwidth]{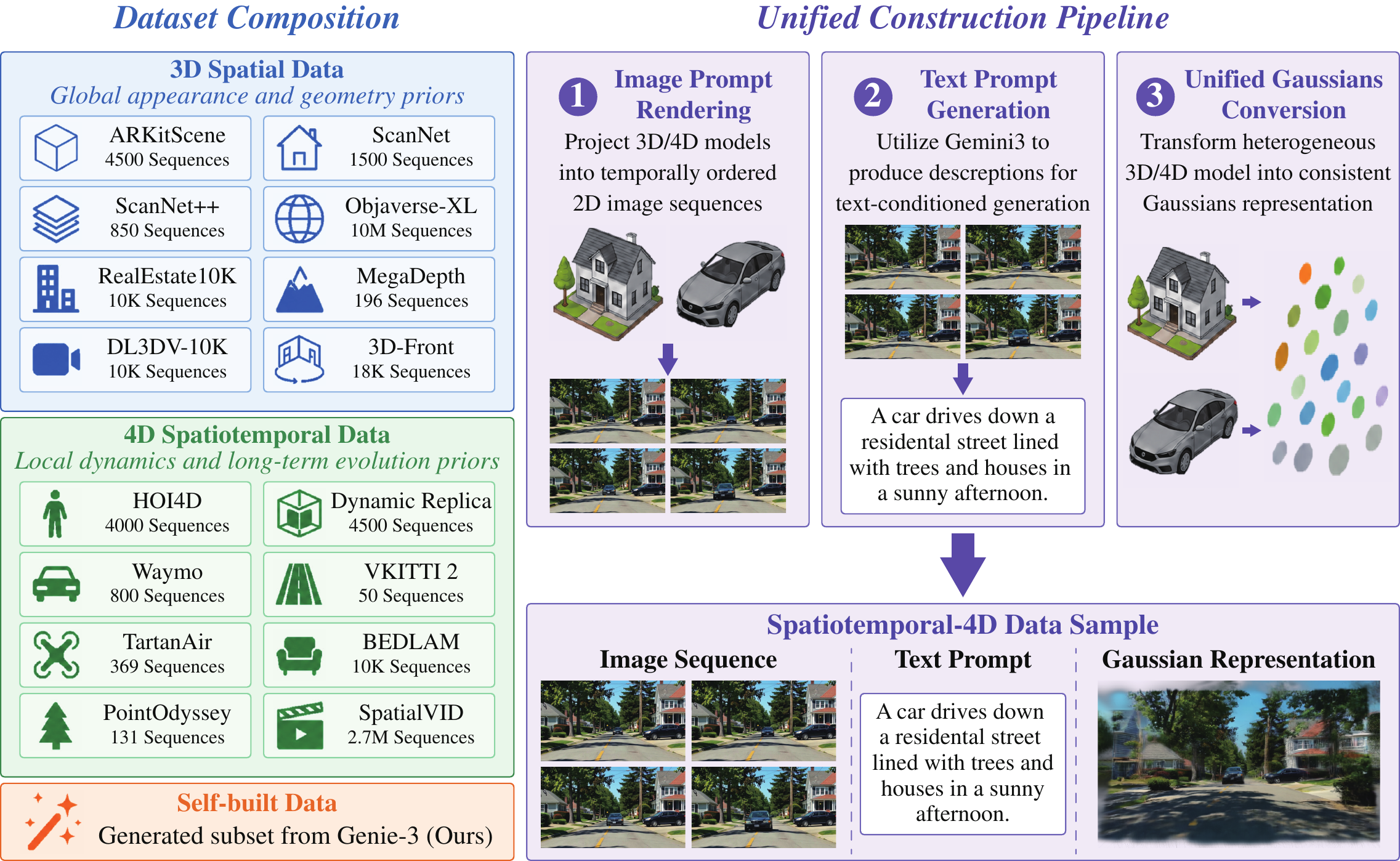}
	\caption{
		Composition and construction pipeline of ST-4D.
	} 
	\label{fig:datasets}
\end{wrapfigure}

As shown in Fig. \ref{fig:datasets}, we propose \textbf{ST-4D}, a comprehensive spatiotemporal 4D dataset that integrates public sources and self-built data generated from Genie-3 \cite{bruce2024genie}, which is designed to support cognition-based 4D generation in a unified formulation. It aggregates diverse 3D spatial datasets \cite{baruch2021arkitscenes,dai2017scannet,yeshwanth2023scannet++,deitke2023objaverse,zhou2018stereo,li2018megadepth,ling2024dl3dv,fu20213d} and 4D spatiotemporal datasets \cite{liu2022hoi4d,karaev2023dynamicstereo,sun2020scalability,cabon2020virtual,wang2020tartanair,black2023bedlam,zheng2023pointodyssey,wang2025spatialvid}, covering global appearance, local dynamics, and long-term dynamic evolution, which provide essential priors for our whole framework. To unify heterogeneous sources, we standardize the data through three steps: \emph{Image prompt rendering} projects 3D/4D assets into temporally ordered 2D sequences as visual prompts. \emph{Text prompt generation} employs Gemini3 to produce semantic descriptions for text-conditioned generation. \emph{Unified Gaussian conversion} transforms diverse formats into consistent Gaussian representations for efficient differentiable rendering and diffusion training. With this unified design, ST-4D serves as a training foundation that connects spatiotemporal representation learning, cognition graph construction, world model reasoning, and cognition-conditioned 4D generation.

\subsection{Training Pipeline}




We employ a progressive three-stage strategy to incrementally develop capabilities from foundational understanding to consistent 4D generation:

\textbf{Stage 1: Foundational Cognition-Evolution Learning.} To construct the feature foundation, we fix the action encoder and utilize Wan2.1 \cite{wan2025wan} as the state encoder to optimize the predictor. Subsequently, we freeze the predictor and train the state encoder with spatiotemporal cognition graph as input. This provides structurally consistent representations and facilitates long-term state evolution.

\textbf{Stage 2: Pixel Appearance Alignment.} To instill high-fidelity visual perception and generative priors, we pre-train the model on large-scale video datasets using Score Distillation Sampling (SDS). Leveraging massive video data and multi-view diffusion models, this warm-up phase enables the framework to generate visually plausible scenes.

\textbf{Stage 3: Spatiotemporal Perception Refinement.} Finally, we conduct global fine-tuning using 3D/4D point cloud data to refine spatiotemporal perception. This comprehensive refinement ensures the generated 4D scenes maintain plausible global appearance and coherent local dynamics.

\begin{table}[t]
	\centering
	\setlength{\abovecaptionskip}{0.0cm}
	\setlength{\belowcaptionskip}{0.0cm}
	\captionsetup{font={scriptsize}}
	\addvbuffer[0pt -6pt]{	
		\begin{minipage}[t]{0.552\textwidth}
			\vspace{0pt} 
			\centering
			\caption{Text-to-3D comparison on T3Bench.}
			\label{tab1}
			\setlength{\tabcolsep}{3.0pt}
			\renewcommand{\arraystretch}{0.8}
			\resizebox{\linewidth}{!}{
				\begin{tabular}{clcccc}
					\toprule
					\multicolumn{2}{c}{\multirow{2}{*}{Method}} & \multicolumn{4}{c}{T3Bench \cite{he2023t}} \\
					\cmidrule(lr){3-6}
					& & Single Obj.$\uparrow$ & Single w/ Surr.$\uparrow$ & Multi Obj.$\uparrow$ & Average$\uparrow$ \\
					\midrule
					\multirow{6}{*}{3D Generation} & DREAMFUSION \cite{poole2022dreamfusion} & 24.4 & 19.8 & 11.7 & 18.7 \\
					& Magic3D \cite{lin2023magic3d} & 37 & 35.4 & 25.7 & 32.7 \\
					& SJC \cite{wang2023score} & 24.7 & 19.8 & 11.7 & 18.7 \\
					& ProlificDreamer \cite{wang2023prolificdreamer} & 49.4 & 44.8 & 35.8 & 43.3 \\
					& GaussianDreamer \cite{yi2024gaussiandreamer} & 54 & 48.6 & 34.5 & 45.7 \\
					& Cog2Gen3D \cite{wang2026cog2gen3d} & 58.3 & 57.9 & 53.6 & 56.6 \\
					\midrule
					\multirow{3}{*}{4D Generation} & 4Dfy \cite{bahmani20244d} & 52.6 & 45.1 & 36.1 & 44.6 \\
					& Dream-in-4D \cite{zheng2024unified} & 55.2 & 50.3 & 42.4 & 49.3 \\
					& \textbf{Ours} & \textbf{58.9} & \textbf{58.7} & \textbf{55.2} & \textbf{57.6} \\
					\bottomrule
				\end{tabular}
			}
		\end{minipage}\hfill 
		\begin{minipage}[t]{0.448\textwidth}
			\vspace{0pt} 
			\centering
			\caption{Image-to-3D comparison on 3D-Front.}
			\label{tab2}
			\setlength{\tabcolsep}{2.5pt}
			\renewcommand{\arraystretch}{0.8}
			\resizebox{\linewidth}{!}{
				\begin{tabular}{clccc}
					\toprule
					\multicolumn{2}{c}{\multirow{2}{*}{Method}} & \multicolumn{3}{c}{3D-Front \cite{fu20213d}} \\
					\cmidrule(lr){3-5}
					& & Chamfer Distance $\downarrow$ & F-Score $\uparrow$ & IoU-B $\uparrow$ \\
					\midrule
					\multirow{5}{*}{3D Generation} & LucidDreamer \cite{chung2023luciddreamer} & 0.138 & 39.99 & 0.299 \\
					& Gen3DSR \cite{ardelean2025gen3dsr} & 0.123 & 40.07 & 0.363 \\
					& EchoScene \cite{zhai2024echoscene} & 0.105 & 45.62 & 0.458 \\
					& Layout2Scene \cite{chen2025layout2scene} & 0.094 & 48.36 & 0.492 \\
					& Cog2Gen3D \cite{wang2026cog2gen3d} & 0.063 & 58.43 & 0.682 \\
					\midrule
					\multirow{4}{*}{4D Generation} & Free4D \cite{liu2025free4d} & 0.114 & 42.85 & 0.412 \\
					& NeoVerse \cite{yang2026neoverse} & 0.099 & 47.14 & 0.475 \\
					& WorldFM  \cite{team2026inspatio} & 0.078 & 53.46 & 0.587 \\
					& \textbf{Ours} & \textbf{0.051} & \textbf{62.15} & \textbf{0.713} \\
					\bottomrule
				\end{tabular}
			}
		\end{minipage}
	}
\end{table}

\begin{figure}[t]
	\vspace{-3pt}
	\centering
	\setlength{\abovecaptionskip}{0.1cm} 
	\setlength{\belowcaptionskip}{-0.3cm} 
	
	\includegraphics[scale=0.103]{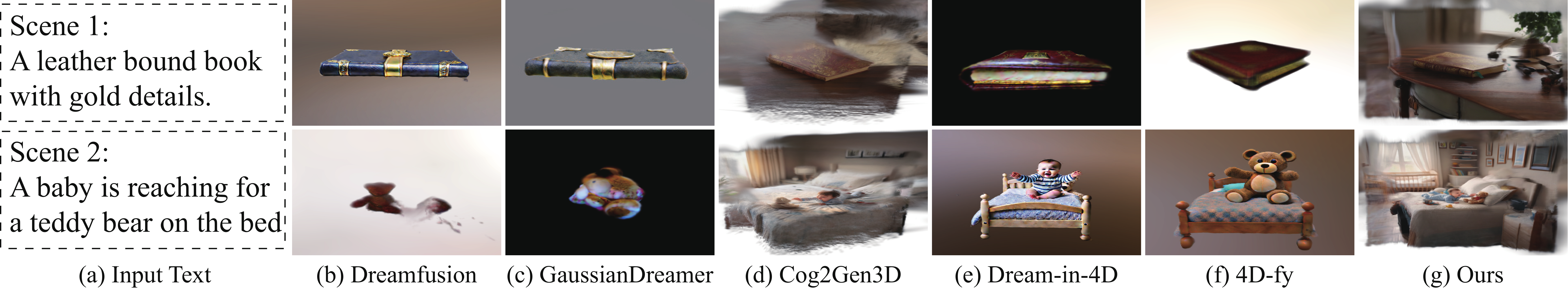}
	\caption{Visualization of Text-to-3D generation on the T3Bench dataset.}
	\label{fig6}
	
	\vspace{10pt} 
	
	\includegraphics[scale=0.065]{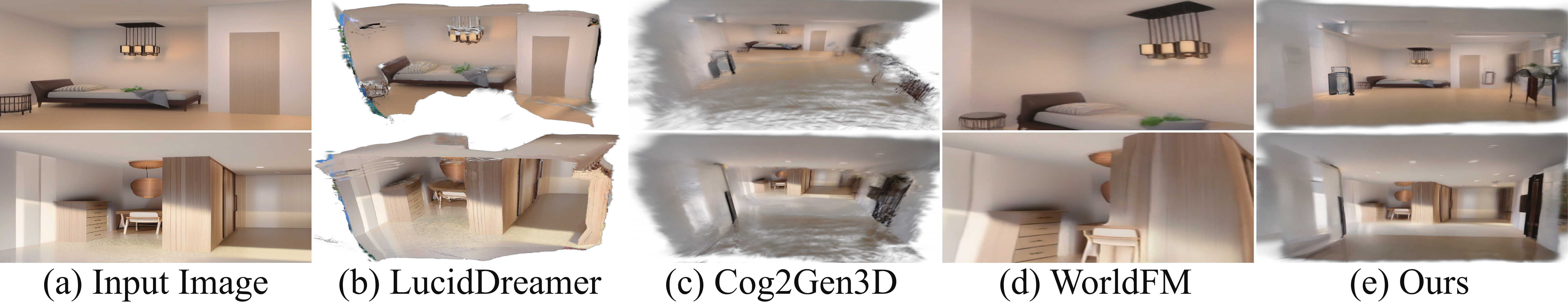}
	\caption{Visualization of image-to-3D generation on the 3D-Front dataset.}
	\label{fig7}
	\vspace{-6pt}
\end{figure}

\section{Experiments}
\label{sec5}

\subsection{Implementation Details}
\label{sec5.1}

For semantic-global-local graph construction, we extract $n=256$ nodes per graph to balance representational capacity and computational efficiency. The latent diffusion model employs 1000 noise timesteps for flow-matching and 50 inference steps during sampling. We optimize the network using AdamW with an initial learning rate of $1 \times 10^{-4}$, a 0.01 weight decay, linear warmup, and cosine annealing. We set the internal loss weights in Eq. (\ref{eq6}) and Eq. (\ref{eq8}) to $\alpha=0.1$, $\beta=0.5$, and $\gamma=1.0$. During the final joint fine-tuning stage, the total loss weights in Eq. (\ref{eq9}) are configured as $\lambda_1=1.0$, $\lambda_2=0.1$, and $\lambda_3=0.5$. Training and evaluation utilize 8 NVIDIA A800 GPUs.

\begin{table}[t]
	\centering
	\setlength{\abovecaptionskip}{0.0cm}
	\setlength{\belowcaptionskip}{0.0cm}
	\captionsetup{font={scriptsize}}
	\addvbuffer[0pt -9pt]{	
		\begin{minipage}[t]{0.57\textwidth}
			\vspace{0pt} 
			\centering
			\caption{Text-to-4D evaluation Vbench and User Study.}
			\label{tab3}
			\setlength{\tabcolsep}{2.0pt}
			\renewcommand{\arraystretch}{0.9}
			\resizebox{\linewidth}{!}{
				\begin{tabular}{lcccccccc}
					\toprule
					\multirow{2}{*}{Method} & \multicolumn{4}{c}{VBench \cite{huang2024vbench}} & \multicolumn{4}{c}{User Study} \\
					\cmidrule(lr){2-5} \cmidrule(lr){6-9}
					& Text Align$\uparrow$ & Consist.$\uparrow$ & Dyna.$\uparrow$ & Aesth.$\uparrow$ & Align.$\uparrow$ & Qual.$\uparrow$ & Consist.$\uparrow$ & Smooth.$\uparrow$ \\
					\midrule
					AYG \cite{ling2024align} & 22.35 & 89.41 & 36.45 & 43.27 & 6.91 & 7.45 & 7.82 & 7.55 \\
					4Dfy \cite{bahmani20244d} & 25.12 & 91.75 & 52.45 & 52.63 & 7.54 & 8.36 & 8.21 & 8.24 \\
					Dream-in-4D \cite{zheng2024unified} & 24.64 & 90.23 & 53.72 & 53.14 & 7.78 & 8.15 & 8.52 & 8.63 \\
					4Real \cite{yu20244real} & 25.87 & 92.14 & 47.13 & 55.69 & 7.95 & 8.42 & 8.36 & 8.45 \\
					Free4D \cite{liu2025free4d} & 25.71 & 93.56 & 54.27 & 59.42 & 8.12 & 8.71 & 9.04 & 8.55 \\
					\textbf{Ours} & \textbf{36.92} & \textbf{97.43} & \textbf{67.75} & \textbf{69.84} & \textbf{8.79} & \textbf{9.56} & \textbf{9.67} & \textbf{9.28} \\
					\bottomrule
				\end{tabular}
			}
		\end{minipage}\hfill 
		\begin{minipage}[t]{0.43\textwidth}
			\vspace{0pt} 
			\centering
			\caption{Image-to-4D evaluation on Vbench and User Study.}
			\label{tab4}
			\setlength{\tabcolsep}{2.5pt}
			\renewcommand{\arraystretch}{0.8}
			\resizebox{\linewidth}{!}{
				\begin{tabular}{lccccccc}
					\toprule
					\multirow{2}{*}{Method} & \multicolumn{3}{c}{VBench \cite{huang2024vbench}} & \multicolumn{4}{c}{User Study} \\
					\cmidrule(lr){2-4} \cmidrule(lr){5-8}
					& Consis.$\uparrow$ & Dyna.$\uparrow$ & Aesth.$\uparrow$ & Align.$\uparrow$ & Qual.$\uparrow$ & Consist.$\uparrow$ & Smooth.$\uparrow$ \\
					\midrule
					GenXD \cite{zhao2024genxd} & 82.52 & 86.34 & 47.58 & 7.42 & 7.76 & 7.88 & 7.53 \\
					DimensionX \cite{sun2024dimensionx} & 89.14 & 89.73 & 52.39 & 8.21 & 8.18 & 8.57 & 8.49 \\
					Free4D \cite{liu2025free4d} & 92.35 & 94.38 & 56.87 & 8.52 & 8.83 & 9.05 & 8.68 \\
					4DNeX \cite{chen20254dnex} & 85.43 & 90.53 & 49.27 & 7.95 & 8.26 & 8.34 & 8.11 \\
					One4D \cite{mi2025one4d} & 89.62 & 82.74 & 48.56 & 7.84 & 8.45 & 8.12 & 7.95 \\
					NeoVerse \cite{yang2026neoverse} & 91.47 & 92.16 & 51.56 & 8.35 & 8.64 & 8.71 & 8.52 \\
					WorldFM \cite{team2026inspatio} & 93.62 & 95.42 & 54.14 & 8.84 & 9.15 & 8.98 & 8.76 \\
					\textbf{Ours} & \textbf{97.83} & \textbf{99.74} & \textbf{59.37} & \textbf{9.46} & \textbf{9.72} & \textbf{9.85} & \textbf{9.12} \\
					\bottomrule
				\end{tabular}
			}
		\end{minipage}
	}
\end{table}

\begin{figure}[t]
	\centering
	\setlength{\abovecaptionskip}{0.1cm} 
	
	\setlength{\belowcaptionskip}{-0.3cm} 
	\includegraphics[scale=0.105]{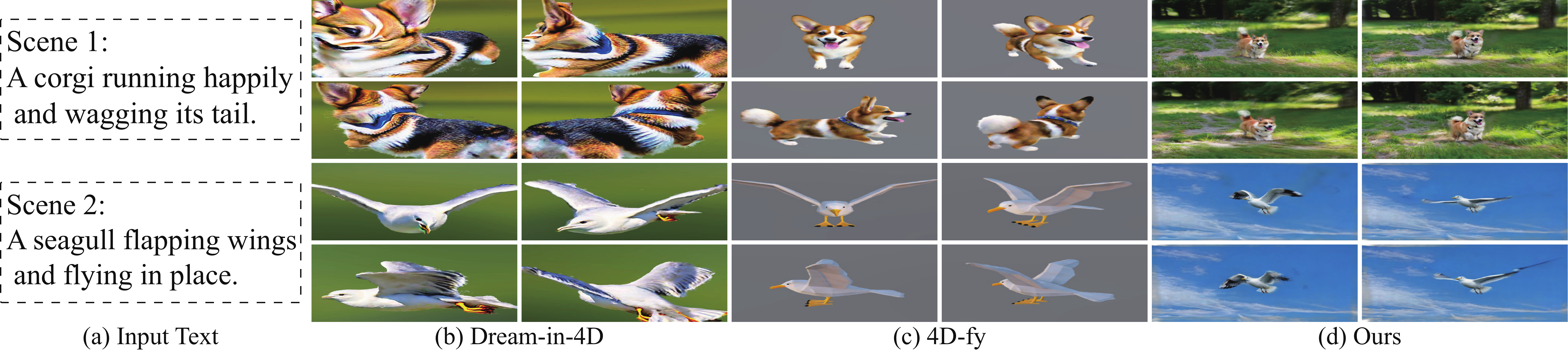}
	\caption{Visualization of text-to-4D generation on the ST-4D dataset.}
	\label{fig8}
	
	\vspace{10pt} 
	
	\setlength{\belowcaptionskip}{-0.3cm} 
	\includegraphics[scale=0.10]{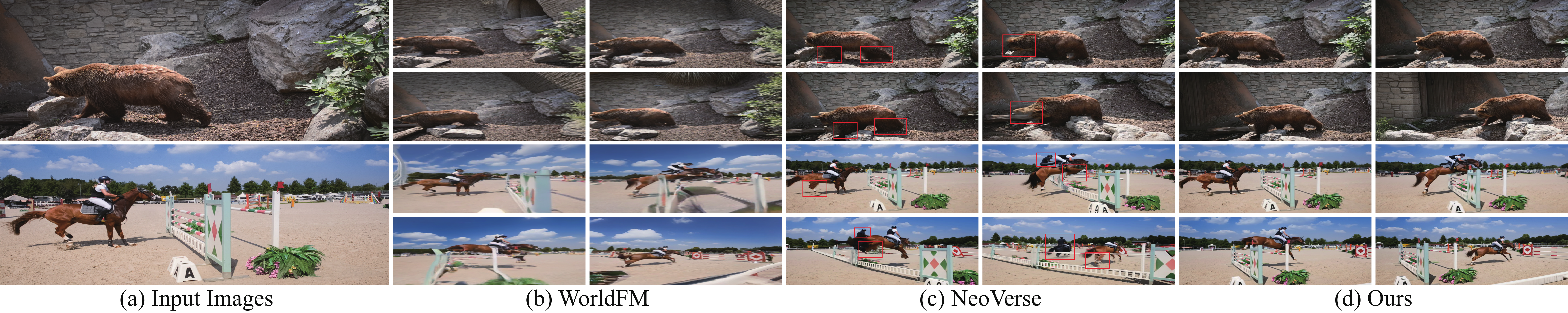}
	\caption{Visualization of image-to-4D generation on the DAVIS \cite{pont20172017} dataset.}
	\label{fig9}
\end{figure}

\subsection{Comparison Experiments}
\textbf{Text-to-3D Generation.} 
We evaluate text-to-3D generation on T3Bench \cite{he2023t} across three evaluation levels against SOTA methods \cite{poole2022dreamfusion,lin2023magic3d,wang2023score,wang2023prolificdreamer,yi2024gaussiandreamer,wang2026cog2gen3d,bahmani20244d,zheng2024unified}. As Table \ref{tab1} and Fig. \ref{fig6} demonstrate, ST-Gen4D achieves superior quantitative scores across all metrics, especially in the challenging multi-object task. Qualitatively, our method consistently generates 3D scenes with detailed appearances and structures, overcoming the blurred appearances and collapsed structures typical of competing methods.

\textbf{Image-to-3D Generation.} 
We further evaluate image-to-3D generation on 3D-Front \cite{fu20213d} against current baselines \cite{liu2025free4d,yang2026neoverse,team2026inspatio,wang2026cog2gen3d,chung2023luciddreamer,ardelean2025gen3dsr,zhai2024echoscene,chen2025layout2scene} using Chamfer Distance, F-Score, and IoU-B. As Table \ref{tab2} and Fig. \ref{fig7} show, our framework significantly outperforms competitors across all quantitative metrics. Qualitatively, our method synthesizes high-fidelity 3D assets with coherent topologies, overcoming the distorted geometries frequent in baseline models.

\textbf{Text-to-4D Generation.} 
We evaluate text-to-4D generation against SOTA methods \cite{ling2024align,bahmani20244d,zheng2024unified,liu2025free4d,yu20244real} using VBench \cite{huang2024vbench} and a user study. As Table \ref{tab3} and Fig. \ref{fig8} demonstrate, our framework secures the highest scores in both automated and human evaluations. Qualitatively, our method synthesizes 4D scenes with seamless state transitions and natural movements, overcoming the severe motion artifacts and temporal flickering typical of competing approaches.

\textbf{Image-to-4D Generation.} 
We evaluate image-to-4D generation against competitive baselines \cite{liu2025free4d,yang2026neoverse,team2026inspatio,zhao2024genxd,sun2024dimensionx,chen20254dnex,mi2025one4d} using VBench \cite{huang2024vbench} and a user study. As Table \ref{tab4} and Fig. \ref{fig9} show, our framework achieves superior quantitative performance across all automated metrics and human preferences. Qualitatively, our method generates highly realistic 4D representations, overcoming the severe appearance degradation and dynamics drift typical of alternative models.

\subsection{Ablation Study}

\textbf{Effectiveness of Multiple Spatiotemporal Representations.} 
To validate our spatiotemporal representations, we ablate the semantic, spatial, and temporal tokens, evaluating three metrics through a user study. As Table \ref{tab5} demonstrates, omitting any token severely degrades its primary metric while compromising other dimensions. This confirms that the synergistic integration of these tokens establishes a robust cognitive foundation for 4D scene synthesis.

\textbf{Superiority of Graph Construction.} 
To validate the graph construction, we replace the structured graph formulation with a direct concatenation of all extracted tokens. Table \ref{tab6} shows that this simple concatenation causes a substantial decline across all VBench metrics, confirming that graph construction is crucial for robust 4D generation.

\textbf{Impact of Separate Global and Local Graphs.} 
To validate the global appearance and local dynamic graphs, we independently ablate each branch. Fig. \ref{fig10} shows that omitting either graph significantly degrades performance in its respective domain. This confirms that the global graph is essential for spatial appearance consistency, and the local graph is vital for temporal dynamic coherence.

\begin{figure}[t!]
	\centering
	\vspace{-10pt}
	
	\captionsetup{type=table} 
	\setlength{\abovecaptionskip}{0.1cm}
	\setlength{\belowcaptionskip}{0cm}
	\setlength{\tabcolsep}{3.0pt}
	\renewcommand{\arraystretch}{0.6}
	\caption{Ablation study of different tokens on user study metrics.}
	\label{tab5}
	\resizebox{\linewidth}{!}{
		\begin{tabular}{cccccc}
			\toprule
			\multicolumn{3}{c}{Settings} & \multicolumn{3}{c}{User Study} \\
			\cmidrule(lr){1-3} \cmidrule(lr){4-6}
			Semantic Tokens & Spatial Tokens & Temporal Tokens & Semantic Fidelity $\uparrow$ & Appearance Plausibility $\uparrow$ & Dynamics Coherence $\uparrow$ \\
			\midrule
			\checkmark & & & 8.2 & 4.3 & 3.5 \\
			& \checkmark & & 4.1 & 8.5 & 3.8 \\
			& & \checkmark & 4.5 & 4.8 & 8.0 \\
			\checkmark & \checkmark & & 8.9 & 9.2 & 4.6 \\
			\checkmark & & \checkmark & 9.0 & 5.2 & 8.6 \\
			& \checkmark & \checkmark & 5.3 & 9.0 & 8.5 \\
			\checkmark & \checkmark & \checkmark & \textbf{9.4} & \textbf{9.7} & \textbf{9.3} \\
			\bottomrule
		\end{tabular}
	}
	
	\vspace{5pt} 
	
	\captionsetup{type=table} 
	\captionsetup{font={scriptsize}}
	\setlength{\abovecaptionskip}{0.1cm}
	\setlength{\belowcaptionskip}{0.0cm}
	
	\begin{minipage}[t]{0.36\textwidth}
		\centering
		\caption{Graph construction ablation.}
		\label{tab6}
		\setlength{\tabcolsep}{3.0pt}
		\renewcommand{\arraystretch}{0.6}
		\resizebox{\linewidth}{!}{
			\begin{tabular}{lccc}
				\toprule
				\multirow{2}{*}{Settings} & \multicolumn{3}{c}{VBench \cite{huang2024vbench}} \\
				\cmidrule(lr){2-4}
				& Consis. $\uparrow$ & Dynam. $\uparrow$ & Aesth. $\uparrow$ \\
				\midrule
				w/o Graph & 86.24 & 87.15 & 46.08 \\
				\textbf{w/. Graph} & \textbf{97.83} & \textbf{99.74} & \textbf{59.37} \\
				\bottomrule
			\end{tabular}
		}
	\end{minipage}\hfill
	\begin{minipage}[t]{0.31\textwidth}
		\centering
		\caption{Semantic encoders discussion.}
		\label{tab7}
		\setlength{\tabcolsep}{3.0pt}
		\renewcommand{\arraystretch}{0.6}
		\resizebox{\linewidth}{!}{
			\begin{tabular}{lccc}
				\toprule
				\multirow{2}{*}{Settings} & \multicolumn{3}{c}{VBench \cite{huang2024vbench}} \\
				\cmidrule(lr){2-4}
				& Consis. $\uparrow$ & Dynam. $\uparrow$ & Aesth. $\uparrow$ \\
				\midrule
				ResNet50 & 88.45 & 90.82 & 49.64 \\
				CLIP ViT-L & 95.62 & 97.10 & 56.21 \\
				\textbf{Wan2.1} & \textbf{97.83} & \textbf{99.74} & \textbf{59.37} \\
				\bottomrule
			\end{tabular}
		}
	\end{minipage}\hfill
	\begin{minipage}[t]{0.32\textwidth}
		\centering
		\caption{Fusion strategy discussion.}
		\label{tab8}
		\setlength{\tabcolsep}{3.0pt}
		\renewcommand{\arraystretch}{0.6}
		\resizebox{\linewidth}{!}{
			\begin{tabular}{lccc}
				\toprule
				\multirow{2}{*}{Settings} & \multicolumn{3}{c}{VBench \cite{huang2024vbench}} \\
				\cmidrule(lr){2-4}
				& Consis. $\uparrow$ & Dynam. $\uparrow$ & Aesth. $\uparrow$ \\
				\midrule
				Concat. & 90.10 & 91.85 & 52.12 \\
				Weighted & 93.85 & 95.60 & 55.20 \\
				\textbf{Sem-Bridged} & \textbf{97.83} & \textbf{99.74} & \textbf{59.37} \\
				\bottomrule
			\end{tabular}
		}
	\end{minipage}
	
	\vspace{5pt} 
	
	\captionsetup{type=figure} 
	\captionsetup{font={normalsize}}
	\centering
	\setlength{\abovecaptionskip}{0.1cm} 
	\setlength{\belowcaptionskip}{-0.7cm} 
	\includegraphics[scale=0.062]{Figure_10_Discussion-eps-converted-to.pdf}
	\caption{\textbf{Ablation of separate global and local graphs.} The visualization reveal that ablating global or local graphs leads to obvious appearance degradation or motion artifacts. Moreover, the performance metrics significantly declined when ablating either branch.}
	\label{fig10}
	
\end{figure}

\subsection{Discussion}
\label{sec5.4}
\textbf{Influence of Semantic Backbone.} To verify the effectiveness of our chosen semantic encoder, we replacing the Wan2.1 \cite{wan2025wan} encoder with conventional ResNet50 and CLIP ViT-L. Table \ref{tab7} demonstrates that introducing the Wan2.1 encoder provides superior semantic representations.

\textbf{Effectiveness of Semantic-Bridged Fusion Strategy.} We replace our semantic-bridged fusion strategy with basic concatenation and weight fusion baselines to verify its effectiveness. Table \ref{tab8} highlights that semantic-bridged fusion facilitate plausible 4D generation.

\textbf{Limitations.} 
Despite its robust 4D generation capabilities, our framework faces challenges with extremely long-duration dynamics. The autoregressive inference of spatiotemporal cognition graphs allows minor early topological inaccuracies to accumulate, potentially causing structural degradation over extended sequences. To mitigate error propagation and extend the generation horizon, future work will explore bidirectional temporal constraints and global trajectory refinement modules.

\section{Conclusion}
In this work, we propose ST-Gen4D, a cognition-guided framework that embeds spatiotemporal cognition into a world model for 4D generation. By integrating disentangled spatial appearance and temporal dynamics representations, our method effectively resolves global apparent collapse and local dynamic artifacts prevalent in existing 4D generation models. Specifically, we construct a spatiotemporal cognition graph from foundational multimodal tokens to capture complex topological dependencies. An action-conditioned world model then autoregressively infers future cognition states, which effctively steer a latent flow-matching diffusion process for 4D Gaussian generation. Extensive experiments across 3D and 4D generation tasks demonstrate the superiority of our approach, affirming its exceptional capability in modeling plausible spatial appearance and coherent temporal dynamics.

\bibliographystyle{splncs04}
\bibliography{main}

\end{document}